\documentclass{article}
\pdfoutput=1
\usepackage[accepted]{icml2017}
\usepackage{tikz}
\usetikzlibrary{fit,positioning}
\usepackage{diagbox}
\usepackage{diagbox}
\usepackage{bbm}
\usepackage{dsfont}
\usepackage{times}
\usepackage{graphicx} %
\usepackage{subfigure}

\usepackage{natbib}

\usepackage{algorithm}
\usepackage{algorithmic}

\usepackage{graphicx}
\usepackage{amsmath,amsthm,amssymb,bm} %
\usepackage{amsfonts}
\usepackage{mathrsfs}
\usepackage{subfigure}
\usepackage{xspace}
\usepackage{array}
\usepackage{enumerate}

\usepackage{algorithmic}
\usepackage{stmaryrd}
\usepackage{appendix}
\usepackage{wrapfig}
\numberwithin{equation}{section}

\newcommand{\calr}{\mathcal{R}} %
      \newcommand{\cg}{\mathcal{G}}         \newcommand{\cp}{\mathcal{P}}       \newcommand{\cx}{\mathcal{X}}  

\theoremstyle{plain}

\theoremstyle{definition}

\theoremstyle{remark}

\usepackage{color}

\usepackage{hyperref}
\usepackage{graphicx} %
\usepackage{booktabs} 
\usepackage{amssymb}
\usepackage{times}
\usepackage{graphicx}
\usepackage{color}
\usepackage{url}
\usepackage{multicol}

\providecommand{\dif}{\mathop{}\!\mathrm d}

\providecommand{\hide}[1]{}

\usepackage{amsmath,amsfonts}
\usepackage{amsopn,amssymb,amsthm,thmtools,thm-restate}

\theoremstyle{plain}

\theoremstyle{definition}

\theoremstyle{remark}

\usepackage{bm} %
\usepackage{algorithm}%
\newcommand{\vct}[1]{\boldsymbol{#1}} %
\newcommand{\mat}[1]{\boldsymbol{#1}} %
\newcommand{\T}{^{\textrm T}} %

\newcommand{\ProbOpr}[1]{\mathbb{#1}}
\newcommand{\expect}[2]{%
\ifthenelse{\equal{#2}{}}{\ProbOpr{E}_{#1}}
{\ifthenelse{\equal{#1}{}}{\ProbOpr{E}\left[#2\right]}{\ProbOpr{E}_{#1}\left[#2\right]}}} %
\newcommand{\var}[2]{%
\ifthenelse{\equal{#2}{}}{\ProbOpr{VAR}_{#1}}
{\ifthenelse{\equal{#1}{}}{\ProbOpr{VAR}\left[#2\right]}{\ProbOpr{VAR}_{#1}\left[#2\right]}}} %
\DeclareMathOperator{\argmax}{arg\,max}

\newcommand{\vx}{{\vct{x}}}
\newcommand{\vy}{\vct{y}}

\newcommand{\vk}{\vct{k}}

\newcommand{\mI}{\mat{I}}
\newcommand{\mK}{\mat{K}}

\icmltitlerunning{Batched High-dimensional Bayesian Optimization via Structural Kernel Learning}

\begin{document}
\twocolumn[
\icmltitle{Batched High-dimensional Bayesian Optimization \\via Structural Kernel Learning}
\icmlsetsymbol{equal}{*}

\begin{icmlauthorlist}
\icmlauthor{Zi Wang}{equal,mit}
\icmlauthor{Chengtao Li}{equal,mit}
\icmlauthor{Stefanie Jegelka}{mit}
\icmlauthor{Pushmeet Kohli}{deep}
\end{icmlauthorlist}

\icmlaffiliation{mit}{Computer Science and Artificial Intelligence Laboratory, Massachusetts Institute of Technology, Massachusetts, USA}
\icmlaffiliation{deep}{DeepMind, London, UK}

\icmlcorrespondingauthor{Zi Wang}{ziw@csail.mit.edu}
\icmlcorrespondingauthor{Chengtao Li}{ctli@mit.edu}
\icmlcorrespondingauthor{Stefanie Jegelka}{stefje@csail.mit.edu}
\icmlcorrespondingauthor{Pushmeet Kohli}{pushmeet@google.com}

\icmlkeywords{Bayesian optimization, Gaussian process, batched Bayesian optimization, determinantal point processes, additive Gaussian process}

\vskip 0.3in
]
\printAffiliationsAndNotice{\icmlEqualContribution} %

\begin{abstract}
Optimization of high-dimensional black-box functions is an extremely challenging problem. While Bayesian optimization has emerged as a popular approach for optimizing black-box functions, its applicability has been limited to low-dimensional problems due to its computational and statistical challenges arising from high-dimensional settings. In this paper, we propose to tackle these challenges by (1) assuming a latent additive structure in the function and inferring it properly for more efficient and effective BO, and (2) performing multiple evaluations in parallel to reduce the number of iterations required by the method. Our novel approach learns the latent structure with Gibbs sampling and constructs batched queries using determinantal point processes. Experimental validations on both synthetic and real-world functions demonstrate that the proposed method outperforms the existing state-of-the-art approaches.

\end{abstract}

\section{Introduction}
Optimization is one of the fundamental pillars of modern machine learning. Considering that most modern machine learning methods involve the solution of some optimization problem, it is not surprising that many recent breakthroughs in this area have been on the back of more effective techniques for optimization. A case in point is deep learning, whose rise has been mirrored by the development of numerous techniques like batch normalization.

While modern algorithms have been shown to be very effective for convex optimization problems defined over continuous domains, the same cannot be stated for non-convex optimization, which has generally been dominated by stochastic techniques. During the last decade, Bayesian optimization has emerged as a popular approach for optimizing black-box functions. However, its applicability is limited to low-dimensional problems because of computational and statistical challenges that arise from optimization in high-dimensional settings.

In the past, these two problems have been addressed by assuming a simpler underlying structure of the black-box function. For instance, \citet{djolonga2013high} assume that the function being optimized has a low-dimensional effective subspace, and learn this subspace via low-rank matrix recovery. Similarly, \citet{kandasamy2015high} assume additive structure of the function where different constituent functions operate on disjoint low-dimensional subspaces. The subspace decomposition can be partially optimized by searching possible decompositions and choosing the one with the highest GP marginal likelihood (treating the decomposition as a hyper-parameter of the GP). Fully optimizing the decomposition is, however, intractable. \citet{li2016high} extended \cite{kandasamy2015high} to functions with a projected-additive structure, and approximate the projective matrix via projection pursuit with the assumption that the projected subspaces have the same and known dimensions. The aforementioned approaches share the computational challenge of learning the groups of decomposed subspaces without assuming the dimensions of the subspaces are known. Both~\cite{kandasamy2015high} and subsequently~\cite{li2016high} adapt the decomposition by maximizing the GP marginal likelihood every certain number of iterations. However, such maximization is computationally intractable due to the combinatorial nature of the partitions of the feature space, which forces prior work to adopt randomized search heuristics.

In this paper, we develop a new formulation of Bayesian optimization specialized for high dimensions. One of the key contributions of this work is a new formulation that interprets prior work on high-dimensional Bayesian optimization (HDBO) through the lens of structured kernels, and places a prior on the kernel structure. %
Thereby, our formulation enables simultaneous learning of the decomposition of the function domain.

Prior work on latent decomposition of the feature space considers the setting where exploration/evaluation is performed once at a time. This approach makes Bayesian optimization time-consuming for problems where a large number of function evaluations need to be made, which is the case for high dimensional problems. To overcome this restriction, we extend our approach to a
batched version that allows
multiple function evaluations to be performed in parallel~\cite{desautels2014parallelizing,gonzalez2016batch,kathuria2016batched}. Our second contribution is an approach to select the batch of evaluations for structured kernel learning-based 
HDBO.

\hide{
\section{Introduction}

Bayesian optimization has been a popular approach for optimizing black-box functions, but usually the applications are limited to low-dimensional problems because of the statistical problem of function estimation and computational issues involving optimizing the acquisition functions in high dimensions.

In the past, these two problems have been addressed by assuming a simpler underlying structure of the black-box function. \cite{djolonga2013high} assumes the function has a low-dimensional effective subspace, and learn this subspace via low-rank matrix recovery. \cite{kandasamy2015high} assumes additive structure of the function whose components are effective on disjoint low-dimensional subspaces. The subspace decomposition can be partially optimized by selecting random decompositions and choose the one with the highest GP marginal likelihood (treating the decomposition as a hyper parameter of the GP). Fully optimizing the decomposition is, however, intractable. \cite{li2016high} further extended \cite{kandasamy2015high} to functions that have a projected-additive structure, and approximate the projective matrix via projective pursuit. It was assumed that the projected subspaces have the same and known dimensions.

A key question not answered well in the past work is how to learn the groups of decomposed subspaces without assuming the dimension of the subspace is known. We can view this question in a Bayesian way. Assume the dimension of the input is $D$ and the number of groups is $K$. We generate a function via 1) draw group mixing proportion $\theta\sim \textsc{Dir}(\alpha)$; 2) for each dimension $j$, choose a group assignment $z_j \sim \textsc{Multi}(\theta)$. Then function can be decomposed as $f(x) =\sum_{i=1}^K f_i(x^{A_i})$, where $A_{i} = \{{j:\mathbbm{1}(z_j=i)}\}$ is the set of active dimensions for function $i$.

To use the group structure, we can either learn it once before any BO procedures and fix it during BO, or adapt it when we get more observations and do joint group structure learning and BO.
}

\paragraph{Other Related Work.}
 In the past half century, a series of different acquisition functions was developed for sequential BO in relatively low dimensions~\citep{kushner1964,mockus1974, srinivas2012information, hennig2012,hernandez2014predictive,kawaguchi2015bayesian,wang2016est,kawaguchi2016global,wang2017maxvalue}. More recent developments address high dimensional BO by making assumptions on the latent structure of the function to be optimized, such as low-dimensional structure~\citep{wang2016bayesian,djolonga2013high} or additive structure of the function~\citep{li2016high,kandasamy2015high}. \citet{duvenaud13} explicitly search over kernel structures.

While the aforementioned methods are sequential in nature, the growth of computing power has motivated settings where at once a \emph{batch} of points is selected for observation
\citep{contal2013parallel, desautels2014parallelizing,gonzalez2016batch,snoek2012practical,wang17}. For example, the UCB-PE algorithm \citep{contal2013parallel}
exploits that the posterior variance of a Gaussian Process is independent of the function mean. It greedily selects points with the highest posterior variance, and is able to update the variances without observations in between selections.
Similarly, B-UCB~\citep{desautels2014parallelizing} greedily chooses points with the highest UCB score computed via the out-dated function mean but up-to-date function variances. However, these methods may be too greedy in their selection, resulting in points that lie far from an optimum. More recently, \citet{kathuria2016batched} tries to resolve this issue by sampling the batch via a diversity-promoting distribution for better randomized exploration, while~\citet{wang17} quantifies the goodness of the batch with a submodular surrogate function that trades off quality and diversity.

\section{Background}
\label{sec:background}

Let $f:\cx\to\mathbb{R}$ be an unknown function and we aim to optimize it over a compact set $\cx\subseteq \mathbb R^{D}$. Within as few function evaluations as possible, we want to find
\begin{align*}
f(x^*) = \max_{x\in\cx} f(x).
\end{align*}
Following~\cite{kandasamy2015high}, we assume a latent decomposition of the feature dimensions $[D]=\{1,\ldots,D\}$ into disjoint subspaces, namely, $\bigcup_{m=1}^M A_m = [D]$ and $A_i\cap A_j = \emptyset$ for all $i\ne j$, $i,j\in[D]$. Further, $f$ can be decomposed into the following additive form:
\begin{align*}
f(x) &= \sum_{m\in[M]} f_m(x^{A_m}).
\end{align*}
To make the problem tractable, we assume that each $f_m$ is drawn independently from $\cg\cp(0,k^{(m)})$ for all $m\in[M]$. The resulting $f$ will also be a sample from a GP: $f\sim\cg\cp(\mu, k)$, where the priors are
$\mu(x) = \sum_{m\in[M]}\mu_m(x^{A_m})$ and 
$k(x, x') = \sum_{m\in[M]}k^{(m)}(x^{A_m}, {x'}^{A_m})$.
Let $\mathcal D_{n} = \{(x_t, y_t)\}_{t=1}^n$ be the data we observed from $f$, where $y_t\sim\mathcal N(f(x_t),\sigma)$. The log data likelihood for $\mathcal D_{n}$ is 
\begin{align} \label{eq:likelihood}
  &\log p(\mathcal D_n|\{k^{(m)}, A_m\}_{m\in[M]})\\
  &= -\frac12 (\vy\T (\mK_n+\sigma^2 \mI)^{-1}\vy \nonumber
  +\log |\mK_n+\sigma^2 \mI| + n\log 2\pi)
\end{align}
where $\mK_n =\left[ \sum_{m=1}^M k^{(m)}(x_i^{A_m},x_j^{A_m})\right]_{i\leq n,j\leq n}$ is the gram matrix associated with $\mathcal D_n$, and $\vy = [y_t]_{t\leq n}$ are the concatenated observed function values. Conditioned on the observations $\mathcal D_n$, we can infer the posterior mean and covariance function of the function component $f^{(m)}$ to be
\begin{align*}
&\mu_{n}^{(m)}(x^{A_m})  = \vk^{(m)}_n(x^{A_m})\T(\mK_n+\sigma^2\mI)^{-1}\vy, \\
&k_{n}^{(m)}(x^{A_m}, {x'}^{A_m}) = k^{(m)}(x^{A_m}, {x'}^{A_m}) \\
&\;\;\; - \vk^{(m)}_n(x^{A_m})\T(\mK_n+\sigma^2\mI)^{-1} \vk^{(m)}_n({x'}^{A_m}),
\end{align*}
 where $\vk^{(m)}_n(x^{A_m}) = [k^{(m)}(x_t^{A_m},x^{A_m})]_{t\leq n}$.

We use regret to evaluate the BO algorithms, both in the sequential and the batch selection case. For the sequential selection, let $\tilde r_t = \max_{x\in \cx}f(x) - f(x_{t})$ denote the immediate regret at iteration $t$. We are interested in both the averaged cumulative regret $R_T = \frac{1}{T}\sum_{t} \tilde r_t$ and the simple regret $r_T = \min_{t\leq T} \tilde r_t$ for a total number of $T$ iterations. For batch evaluations, $\tilde r_{t} = \max_{x\in \cx, b\in[B]}f(x) - f(x_{t,b})$ denotes the immediate regret obtained by the batch at iteration $t$. The averaged cumulative regret of the batch setting is $R_{T} = \frac{1}{T}\sum_{t}\tilde r_{t}$, and the simple regret $r_{T} = \min_{t\leq T}  \tilde r_{t}$. We use the averaged cumulative regret in the bandit setting, where each evaluation of the function incurs a cost. If we simply want to optimize the function, we use the simple regret to capture the minimum gap between the best point found and the global optimum of the black-box function $f$. Note that the averaged cumulative regret upper bounds the simple regret.

\hide{
At time step $t$, we select point $\vx_t$ and observe a possibly noisy function evaluation $y_t=f(\vx_t)+\epsilon_t$, where $\epsilon_t$ are i.i.d.\ Gaussian noise $\mathcal N(0,\sigma^2)$.
Given the observations $\mathfrak \mathcal D_t=\{(\vx_\tau,y_\tau)\}_{\tau=1}^{t}$ up to time $t$, we obtain the posterior mean and covariance of the function via the kernel matrix $\mK_t =\left[k(\vx_i,\vx_j)\right]_{\vx_i,\vx_j\in \mathfrak \mathcal D_t}$ and $\vk_t(x) = [k(\vx_i,\vx)]_{\vx_i\in \mathfrak \mathcal D_t}$~\cite{rasmussen2006gaussian}:
$\mu_{t}(\vx) = \vk_t(\vx)\T(\mK_t+\sigma^2\mI)^{-1}\vy_t$, and %
$k_{t}(\vx,\vx') = k(\vx,\vx') - \vk_t(\vx)\T(\mK_t+\sigma^2\mI)^{-1} \vk_t(\vx')$.
The posterior variance is given by $\sigma^2_{t}(\vx) = k_t(\vx,\vx)$. 
}

\section{Learning Additive Kernel Structure}
\label{ssec:learn_partition}

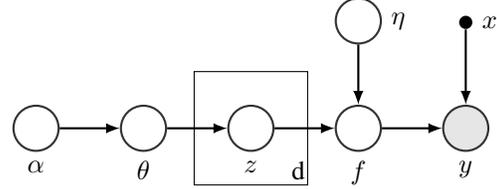
\begin{figure}
\caption{Graphical model for the structured Gaussian process; $\eta$ is the hyperparameter of the GP kernel; $z$ controls the decomposition for the input space.}
\label{fig:addgpgraph}
\centering
\begin{tikzpicture}
\tikzstyle{main}=[circle, minimum size = 6mm, thick, draw =black!80, node distance = 8mm]
\tikzstyle{para}=[circle, minimum size = 5pt, inner sep=0pt]
\tikzstyle{connect}=[-latex, thick]
\tikzstyle{box}=[rectangle, draw=black!100]

  \node[main] (alpha) [label=below:$\alpha$] { };
  \node[main, fill = white!100] (theta) [right=of alpha,label=below:$\theta$] { };
  \node[main] (z) [right=of theta,label=below:$z$] {};
  \node[main] (f) [right=of z,label=below:$f$] { };
  \node[main] (eta) [above=of f,label=right:$\eta$] { };
  \node[main, fill = black!10] (y) [right=of f,label=below:$y$] { };
  \node[para, fill = black!100] (x) [above=of y,label=right:$x$] { };

  \path (alpha) edge [connect] (theta)
        (theta) edge [connect] (z)
		(z) edge [connect] (f)
		(eta) edge [connect] (f)
		(f) edge [connect] (y)
		(x) edge [connect] (y);
  \node[rectangle, inner sep=0mm, fit= (z),label=below right:d, xshift=1mm] {};
  \node[rectangle, inner sep=4.4mm,draw=black!100, fit= (z)] {};
\end{tikzpicture}
\end{figure}

We take a Bayesian view on the task of learning the latent structure of the GP kernel. The decomposition of the input space $\cx$ will be learned simultaneously with optimization as more and more data is observed.  %
Our generative model draws mixing proportions $\theta\sim \textsc{Dir}(\alpha)$. Each dimension $j$ is assigned to one out of $M$ groups via the decomposition assignment variable $z_j \sim \textsc{Multi}(\theta)$.
The objective function is then $f(x) =\sum_{m=1}^M f_m(x^{A_m})$, where $A_{m} = \{{j: z_j=m}\}$ is the set of support dimensions for function $f_m$, and each $f_m$ is drawn from a Gaussian Process. Finally, given an input $x$, we observe $y\sim \mathcal N(f(x), \sigma)$. 
Figure~\ref{fig:addgpgraph} illustrates the corresponding graphical model.

Given the observed data $\mathcal D_{n} = \{(x_t, y_t)\}_{t=1}^n$, we obtain a posterior distribution over possible decompositions $z$ (and mixing proportions $\theta$) that we will include later in the BO process: %
\begin{align*}
p(z, \theta \mid \mathcal D_{n};\alpha) \propto p(\mathcal D_n \mid z) p(z \mid \theta) p(\theta;\alpha).
\end{align*}
Marginalizing over $\theta$ yields the posterior distribution of the decomposition assignment
\begin{align*}
& p(z \mid \mathcal D_{n};\alpha) \propto p(\mathcal D_n \mid z) \int p(z \mid \theta) p(\theta;\alpha)\dif \theta \\
&\;\;\;\propto p(\mathcal D_n \mid z) \frac{\Gamma(\sum_m \alpha_m)}{\Gamma(D + \sum_m \alpha_m)}\prod_m\frac{\Gamma(|A_m| + \alpha_m)}{\Gamma( \alpha_m)}
\end{align*}
where $p(\mathcal D_n|z)$ is the data likelihood~\eqref{eq:likelihood} for the additive GP given a fixed structure defined by $z$.%
\hide{
A prediction $f(x')$ for an unseen point $x'$ with the new model marginalizes over decomposition $z$ in the set $\mathcal{Z}$ of all decompositions:
\begin{align*}
  p(f(x')\mid \mathcal D_n; \alpha) = \sum_{z \in \mathcal{Z}} p(f(x')\mid z, \mathcal D_n; \alpha)p(z|\mathcal D_n; \alpha).
\end{align*}
We approximate this sum via Gibbs sampling.}
We learn the posterior distribution for $z$ via Gibbs sampling, choose the decomposition among the samples that achieves the highest data likelihood, and then proceed with BO. The Gibbs sampler repeatedly draws coordinate assignments $z_j$ according to
\begin{align*}
p(z_j = m &\mid z_{\neg j}, \mathcal D_n;\;\alpha) \propto p(\mathcal D_n \mid z) p(z_j\mid z_{\neg j}) \\
&\propto p(\mathcal D_n \mid z) (|A_m| + \alpha_m) \propto e^{\phi_m},
\end{align*}
where
\begin{align*}
&\phi_m = -\frac12 \vy\T (\mK_n^{(z_j=m)}+\sigma^2 \mI)^{-1}\vy\nonumber\\ &\;\;\;-\frac12\log |\mK_n^{(z_j=m)} +\sigma^2 \mI| +
\log (|A_m| + \alpha_m)
\end{align*}
and $\mK_n^{(z_j=m)}$ is the gram matrix associated with the observations $\mathcal D_n$ by setting $z_j = m$. We can use the Gumbel trick to efficiently sample from this categorical distribution. Namely, we sample a vector of i.i.d standard Gumbel variables $\omega_i$ of length $M$, and then choose the sampled decomposition assignment  $z_j= \argmax_{i\leq M} \phi_i + \omega_i$. %

With a Dirichlet process, we could make the model nonparametric and the number $M$ of possible groups in the decomposition infinite. Given that we have a fixed number of input dimension $D$, we set $M=D$ in practice.

\section{Diverse Batch Sampling}\label{sec:batch}

In real-world applications where function evaluations translate into time-intensive experiments, the typical sequential exploration strategy -- observe one function value, update the model, then select the next observation -- is undesirable.  \emph{Batched Bayesian Optimization (BBO)} \cite{azimi10,contal2013parallel,kathuria2016batched} instead selects a batch of $B$ observations to be made in parallel, then the model is updated with all simultaneously.

Extending this scenario to high dimensions, two questions arise: (1) the acquisition function is expensive to optimize and (2), by itself, does not sufficiently account for exploration. The additive kernel structure improves efficiency for (1). For batch selection (2), we need an efficient strategy that enourages observations that are both informative and non-redundant.
Recent work \cite{contal2013parallel,kathuria2016batched} %
 selects a point that maximizes the acquisition function, and adds additional batch points via a diversity criterion.
In high dimensions, this diverse selection becomes expensive. For example, if each dimension has a finite number of possible values\footnote{While we use this discrete categorical domain to illustrate the batch setting, our proposed method is general and is applicable to continuous box-constrained domains.}, the cost of sampling batch points via a \emph{Determinantal Point Process~(DPP)}, as proposed in~\cite{kathuria2016batched}, grows exponentially with the number of dimensions. The same obstacle arises with the approach by \citet{contal2013parallel}, where points are selected greedily. %
Thus, na\"ive adoptions of these approaches in our setting would result in intractable algorithms. Instead, we propose a general approach that explicitly takes advantage of the structured kernel to enable relevant, non-redundant high-dimensional batch selection. 

We describe our approach for a single decomposition sampled from the posterior; it extends to a distribution of decompositions by sampling a set of decompositions from the posterior and then sampling points for each decomposition individually. Given a decomposition $z$, we define a separate Determinantal Point Process (DPP) on each group of $A_m$ dimensions. A set $S$ of points in the subspace $\mathbb{R}^{|A_m|}$ is sampled with probability proportional to $\det(\mathbf{K}^{(m)}_n(S))$, where $\mathbf{K}^{(m)}_n$ is the posterior covariance matrix of the $m$-th group given $n$ observations, and $\mathbf{K}(S)$ is the submatrix of $\mathbf{K}$ with rows and columns indexed by $S$. Assuming the group sizes are upper-bounded by some constant, sampling from each such DPP individually implies an exponential speedup compared to using the full kernel.

\paragraph{Sampling vs.\ Greedy Maximization} The determinant $\det(\mathbf{K}^{(m)}_n(S))$ measures diversity, and hence the DPP assigns higher probability to diverse subsets $S$. An alternative to sampling is to directly maximize the determinant. While this is NP-hard, a greedy strategy gives an approximate solution, and is used in \cite{kathuria2016batched}, and in \cite{contal2013parallel} as Pure Exploration (PE). We too test this strategy in the experiments. In the beginning, if the GP is not approximating the function well, then greedy may perform no better than a stochastic combination of coordinates, as we observe in Fig.~\ref{fig:batch_walker_simple}.

\paragraph{Sample Combination} Now we have chosen a diverse subset $\cx_m = \{x_i^{(m)}\}_{i\in[B-1]} \subset \mathbb{R}^{|A_m|}$ of size $(B-1)$ for each group $A_m$. We need to combine these subspace points to obtain $B-1$ final batch query points in $\mathbb{R}^D$.
A simple way to combine samples from each group is to do it \emph{randomly without replacement}, i.e., we sample one $x_i^{(m)}$ from each $\cx_m$ uniformly randomly without replacement, and combine the parts, one for each $m\in[M]$, to get one sample in $\mathbb{R}^D$. We repeat this procedure until we have $(B-1)$ points. This retains diversity across the batch of samples, since the samples are diverse within each group of features. 

Besides this random combination, we can also combine samples greedily. We define a \emph{quality function} $\psi_t^{(m)}$ for each group $m\in[M]$ at time $t$, and combine samples to maximize this quality function. Concretely, for the first point, we combine the maximizers $x_*^{(m)} = \arg\max_{x^{(m)}\in \cx_m} \psi_t^{(m)}(x^{(m)})$ from each group. We remove those used parts, $\cx_m \leftarrow \cx_m\backslash\{x_*^{(m)}\}$, and repeat the procedure until we have $(B-1)$ samples. In each iteration, the sample achieving the highest quality score gets selected, while diversity is retained.

Both selection strategies can be combined with a wide range of existing quality and acquisition functions. 

\paragraph{Add-UCB-DPP-BBO}
We illustrate the above framework with GP-UCB \cite{srinivas2012information} as both the acquisition and quality functions. 
The Upper Confidence Bound $(f_t^{(m)})^+$ and Lower Confidence Bound $(f_t^{(m)})^-$ with parameter $\beta_t$ for group $m$ at time $t$ are
\begin{align}\label{eq:ucb}
(f_t^{(m)})^+(x) &= \mu_{t-1}^{(m)}(x) + \beta_t^{1/2}\sigma_t^{(m)}(x);\\
(f_t^{(m)})^-(x) &= \mu_{t-1}^{(m)}(x) - \beta_t^{1/2}\sigma_t^{(m)}(x),\nonumber
\end{align}
and combine the expected value $\mu_{t-1}^{(m)}(x)$ of $f_t^{(m)}$ with the uncertainty %
$\beta_t^{1/2}\sigma_t^{(m)}(x)$.
We set both the acquisition function %
and quality function $\psi_t^{(m)}$ to be $(f_t^{(m)})^+$ for group $m$ at time $t$. 

To ensure that we select points with high acquisition function values, we follow~\cite{contal2013parallel,kathuria2016batched} and define a relevance region $\calr_t^{(m)}$ for each group $m$ as %
\begin{align*}
  \calr_t^{(m)} = & \left\{x\in\cx_m \mid \right. \\
  &\; \left. \mu_{t-1}^{(m)}(x) + 2\sqrt{\beta_{t+1}^{(m)}}\sigma_{t-1}^{(m)}(x) \ge (y_t^{(m)})^\bullet\right\},
\end{align*}
where $(y_t^{(m)})^\bullet = \max_{x^{(m)}\in\cx_m} (f_t^{(m)})^-(x^{(m)})$.
We then use $\calr_t^{(m)}$ as the ground set to sample with PE/DPP. %
The full algorithm is shown in the appendix.

\section{Empirical Results}
We empirically evaluate our approach in two parts: First, we verify the effectiveness of using our Gibbs sampling algorithm to learn the additive structure of the unknown function, and then we test our batch BO for high dimensional problems with the Gibbs sampler. Our code is available at \url{https://github.com/zi-w/Structural-Kernel-Learning-for-HDBBO}.
\subsection{Effectiveness of Decomposition Learning}
\label{sec:exp}
\hide{
\begin{table*}[t]
\caption{Empirical posterior of decompositions learned via Gibbs sampling.}
\label{tab:learn_partition}
\centering
\begin{tabular}{l*{6}{c}}
N & $P_1$ & $P_2$ & $P_3$ & $P_4$ & $P_5$\\
\hline
20 & $0.16 \pm 0.06$ & $0.21 \pm 0.06$ & $0.19 \pm 0.09$ & $0.16 \pm 0.07$ & $0.14 \pm 0.06$ & \\
50 & $0.44 \pm 0.28$ & $0.42 \pm 0.38$ & $0.34 \pm 0.24$ & $0.28 \pm 0.26$ & $0.39 \pm 0.27$ & \\
100 & $0.95 \pm 0.09$ & $0.90 \pm 0.19$ & $0.68 \pm 0.33$ & $0.90 \pm 0.27$ & $0.84 \pm 0.21$ & \\
150 & $1.00 \pm 0.01$ & $0.99 \pm 0.03$ & $1.00 \pm 0.00$ & $1.00 \pm 0.00$ & $1.00 \pm 0.00$ & \\
200 & $1.00 \pm 0.00$ & $0.97 \pm 0.08$ & $0.98 \pm 0.07$ & $1.00 \pm 0.00$ & $0.97 \pm 0.08$ & \\
\end{tabular}
\end{table*}
}

We first probe the effectiveness of using the Gibbs sampling method described in Section~\ref{ssec:learn_partition} to learn the decomposition of the input space. More details of  the experiments including sensitivity analysis for $\alpha$ can be found in the appendix.
\begin{figure*}[h!]
\centering
\includegraphics[width=\textwidth]{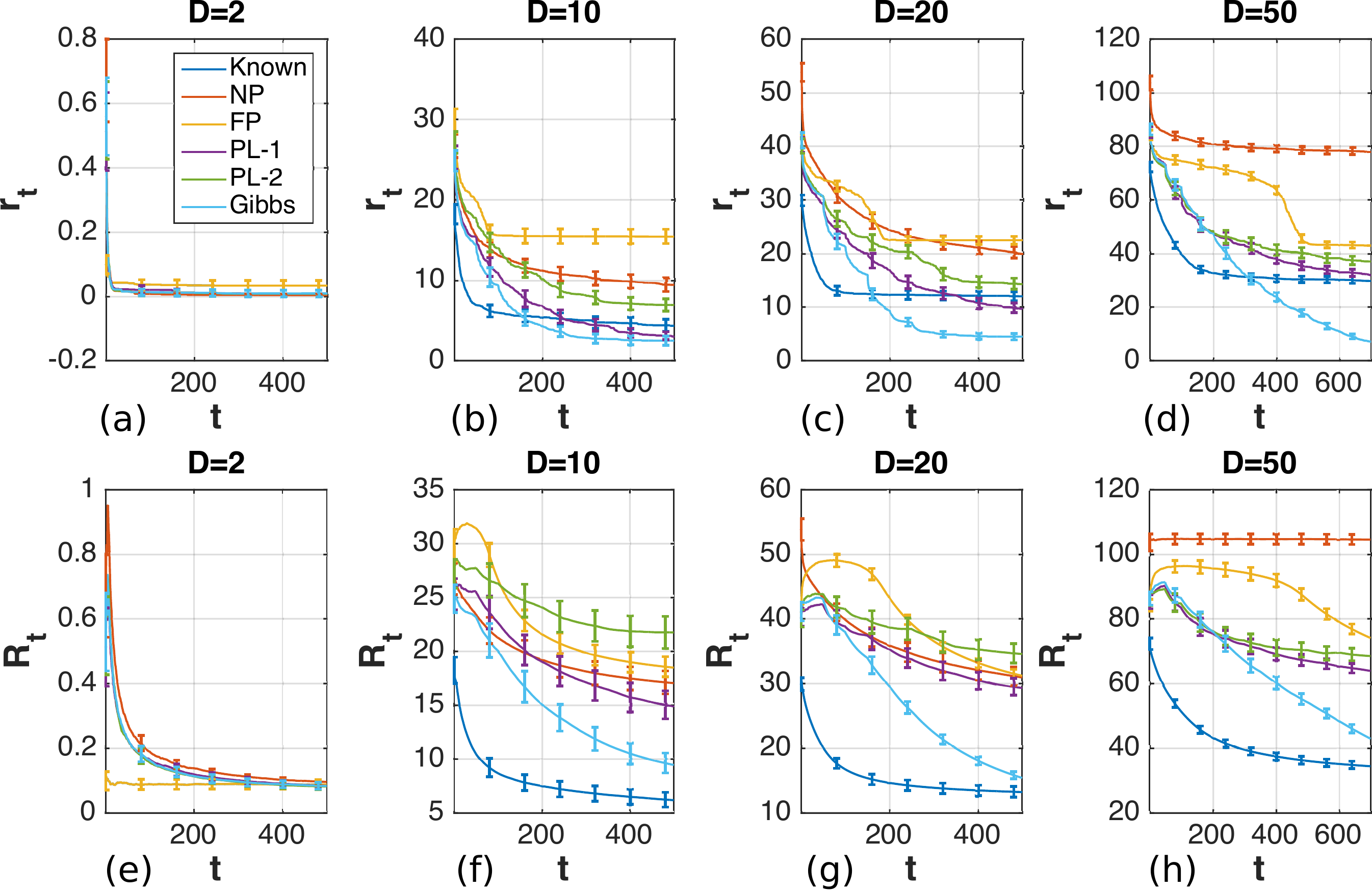}
\caption{The simple regrets ($r_t$) and the averaged cumulative regrets ($R_t$) for setting input space decomposition with \texttt{Known}, \texttt{NP}, \texttt{FP}, \texttt{PL-1}, \texttt{PL-2}, and \texttt{Gibbs} on 2, 10, 20, 50 dimensional synthetic additive functions. \texttt{Gibbs} achieved comparable results to \texttt{Known}. Comparing \texttt{PL-1} and \texttt{PL-2} we can see that sampling more settings of decompositions did help to find a better decomposition. But a more principled way of learning the decomposition using \texttt{Gibbs} can achieve much better performance than \texttt{PL-1} and \texttt{PL-2}. }
\label{fig:synth1}
\end{figure*}

\paragraph{Recovering Decompositions}
First, we sample test functions from a known additive Gaussian Process prior with zero-mean and isotropic Gaussian kernel with bandwidth $=0.1$ and scale $=5$ for each function component. For $D = 2,5,10,20,50,100$ input dimensions, we randomly sample decomposition settings that have at least two groups in the decomposition and at most 3 dimensions in each group.

\vskip -0.2in
\begin{table}[h!]
\caption{Empirical posterior of any two dimensions correctly being grouped together by Gibbs sampling.}
\label{tab:learn_partition_together}
\begin{center}
\begin{scriptsize}
\begin{sc}
\begin{tabular}{l*{5}{c}}
\diagbox[width=3em]{D}{N} & 50 & 150 & 250 & 450 \\
\hline
5 & $0.81 \pm 0.28$ & $0.91 \pm 0.19$ & $1.00 \pm 0.03$ & $1.00 \pm 0.00$  \\
10 & $0.21 \pm 0.13$ & $0.54 \pm 0.25$ & $0.68 \pm 0.25$ &  $0.93 \pm 0.15$  \\
20 & $0.06 \pm 0.06$ & $0.11 \pm 0.08$ & $0.20 \pm 0.12$ &  $0.71 \pm 0.22$  \\
50 & $0.02 \pm 0.03$ & $0.02 \pm 0.02$ & $0.03 \pm 0.03$ & $0.06 \pm 0.04$  \\
100 & $0.01 \pm 0.01$ & $0.01 \pm 0.01$ & $0.01 \pm 0.01$ &  $0.02 \pm 0.02$  \\
\end{tabular}
\end{sc}
\end{scriptsize}
\end{center}
\vskip -0.2in
\end{table}
We set the burn-in period to be 50 iterations, and the total number of iterations for Gibbs sampling to be 100. In Tables~\ref{tab:learn_partition_together} and~\ref{tab:learn_partition_separate}, we show two quantities that are closely related to the learned empirical posterior of the decompositions with different numbers of randomly sampled observed data points ($N$). Table~\ref{tab:learn_partition_together} shows the probability of two dimensions being correctly grouped together by Gibbs sampling in each iteration of Gibbs sampling after the burn-in period, namely, $(\sum_{i<j\leq D}\mathds{1}_{z^{g}_i\equiv z^{g}_j\wedge z_i\equiv z_j})/(\sum_{i<j\leq D}\mathds{1}_{z_i\equiv z_j})$. Table~\ref{tab:learn_partition_separate} reports the probability of two dimensions being correctly separated in each iteration of Gibbs sampling after the burn-in period, namely, $(\sum_{i<j\leq D}\mathds{1}_{z^{g}_i\neq z^{g}_j\wedge z_i\neq z_j})/(\sum_{i<j\leq D}\mathds{1}_{z_i\neq z_j})$. The results %
show that the more data we observe, the more accurate the learned decompositions are. They also suggest that the Gibbs sampling procedure can converge to the ground truth decomposition with enough data for relatively small numbers of dimensions. The higher the dimension, the more data we need to recover the true decomposition.
\vskip -0.2in
\begin{table}[h!]
\caption{Empirical posterior of any two dimensions correctly being separated by Gibbs sampling.}
\label{tab:learn_partition_separate}
\begin{center}
\begin{scriptsize}
\begin{sc}
\begin{tabular}{l*{5}{c}}
\diagbox[width=3em]{D}{N} & 50 & 150 & 250  & 450 \\
\hline
2 & $0.30 \pm 0.46$ & $0.30 \pm 0.46$ & $0.90 \pm 0.30$ &  $1.00 \pm 0.00$ \\
5 & $0.87 \pm 0.17$ & $0.80 \pm 0.27$ & $0.60 \pm 0.32$ &  $0.50 \pm 0.34$  \\
10 & $0.88 \pm 0.05$ & $0.89 \pm 0.06$ & $0.89 \pm 0.07$ &  $0.94 \pm 0.07$  \\
20 & $0.94 \pm 0.02$ & $0.94 \pm 0.02$ & $0.94 \pm 0.02$ &  $0.97 \pm 0.02$  \\
50 & $0.98 \pm 0.00$ & $0.98 \pm 0.00$ & $0.98 \pm 0.01$ &  $0.98 \pm 0.01$  \\
100 & $0.99 \pm 0.00$ & $0.99 \pm 0.00$ & $0.99 \pm 0.00$ &  $0.99 \pm 0.00$  \\

\end{tabular}
\end{sc}
\end{scriptsize}
\end{center}
\vskip -0.2in
\end{table}

\begin{figure}[h]
\centering
\includegraphics[width=0.45\textwidth]{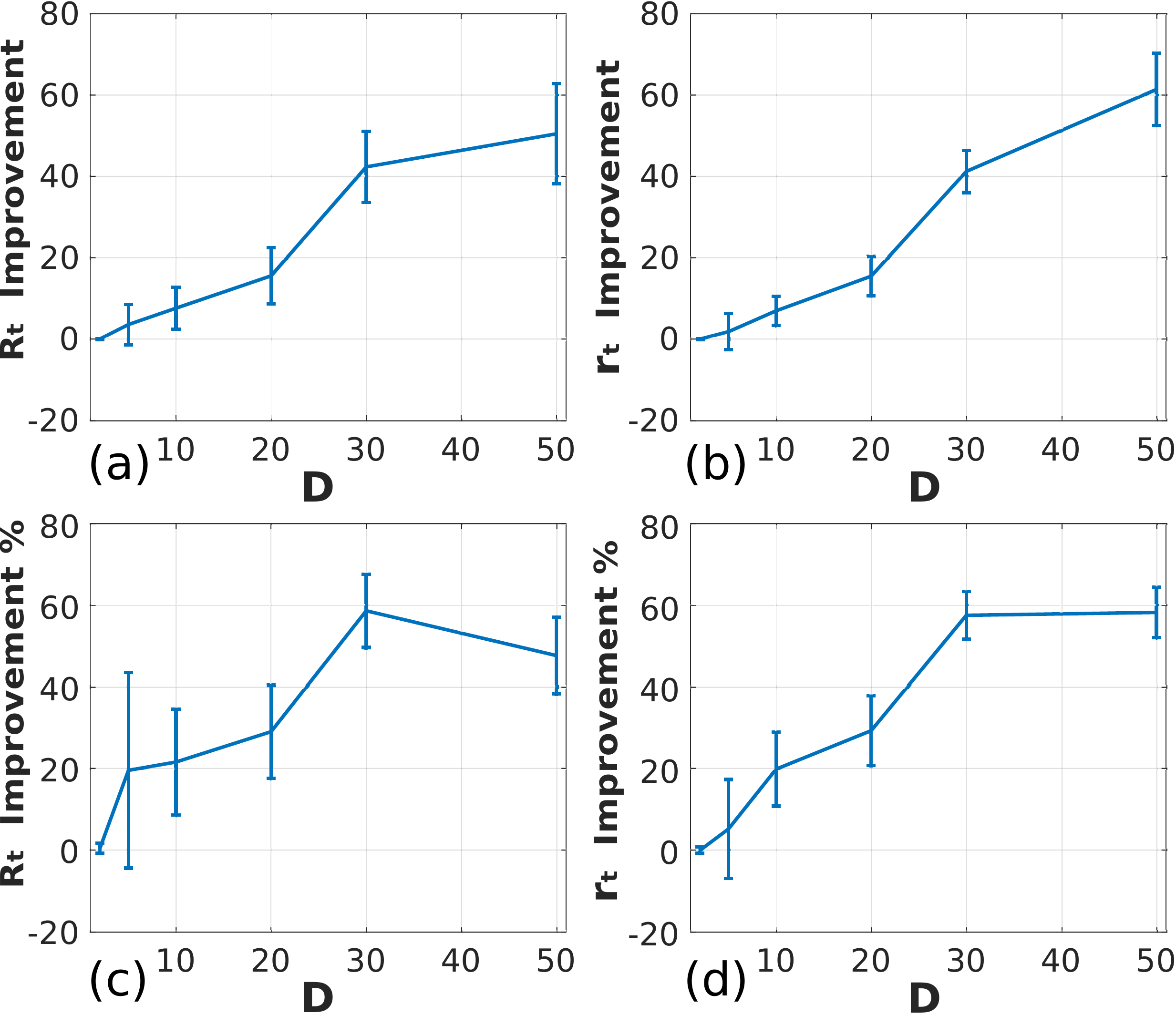}
\caption{Improvement made by learning the decomposition with \texttt{Gibbs} over optimizing without partitions (NP). (a) averaged cumulative regret; (b) simple regret. (c) averaged cumulative regret normalized by function maximum; (d) simple regret normalized by function maximum. Using decompositions learned by \texttt{Gibbs} continues to outperform BO without \texttt{Gibbs}.}
\label{fig:synth2}
\end{figure}

\paragraph{Effectiveness of Learning Decompositions for Bayesian Optimization}
To verify the effectiveness of the learned decomposition for Bayesian optimization,  we tested on 2, 10, 20 and 50 dimensional functions sampled from a zero-mean Add-GP with randomly sampled decomposition settings (at least two groups, at most 3 dimensions in each group) and isotropic Gaussian kernel with bandwidth $=0.1$ and scale $=5$. Each experiment was repeated 50 times. An example of a 2-dimensional function component is shown in the appendix. For Add-GP-UCB, we used $\beta^{(m)}_t=|A_m|\log 2t$ for lower dimensions ($D=2,5,10$), and  $\beta^{(m)}_t=|A_m|\log 2t/5$ for higher dimensions ($D=20,30,50$). We show parts of the results on averaged cumulative regret and simple regret in Fig.~\ref{fig:synth1}, and the rest in the appendix. We compare Add-GP-UCB with known additive structure (\texttt{Known}), no partitions (\texttt{NP}), fully partitioned with one dimension for each group (\texttt{FP})  and the following methods of learning the decomposition: Gibbs sampling (\texttt{Gibbs}), randomly sampling the same number of decompositions sampled by \texttt{Gibbs} and select the one with the highest data likelihood (\texttt{PL-1}), randomly sampling 5 decompositions and selecting the one with the highest data likelihood (\texttt{PL-2}). For the latter two learning methods are referred to as ``partial learning'' in \cite{kandasamy2015high}. The learning of the decomposition is done every 50 iterations.
Fig.~\ref{fig:synth2} shows the improvement of learning decompositions with \texttt{Gibbs} over optimizing without partitions (NP).

Overall, the results show that \texttt{Gibbs} outperforms both of the partial learning methods, and for higher dimensions, \texttt{Gibbs} is sometimes even better than \texttt{Known}. %
Interestingly, similar results can be found in Fig.~3 (c) of~\cite{kandasamy2015high}, where different decompositions than the ground truth may give
better simple regret. 
We conjecture that this is because \texttt{Gibbs} is able to explore more than \texttt{Known}, for
two reasons:
\begin{enumerate}
\item Empirically, \texttt{Gibbs} changes the decompositions across
iterations, especially in the beginning. With fluctuating partitions,
even exploitation leads to moving around, because the supposedly
``good'' points are influenced by the partition. The result is an implicit ``exploration'' effect that is absent with a fixed partition.
\item \texttt{Gibbs} sometimes merges ``true'' parts into larger parts.
The parameter $\beta_t$ in UCB depends on the size of the part, $|A_m|(\log 2t)/5$ (as in~\cite{kandasamy2015high}). Larger parts hence lead to larger $\beta_t$ and hence more exploration.
\end{enumerate}
Of course, more exploration is not always better, but \texttt{Gibbs} was able to find a good balance between exploration and exploitation, which leads to better performance. Our preliminary experiments indicate that one solution to ensure that the ground truth decomposition produces the best result is to tune $\beta_t$. Hyperparameter selection (such as choosing $\beta_t$) for
BO is, however, very challenging and an active topic of research (e.g.~\cite{wang2016est}).

\begin{figure}[h]
\centering
\includegraphics[width=0.4\textwidth]{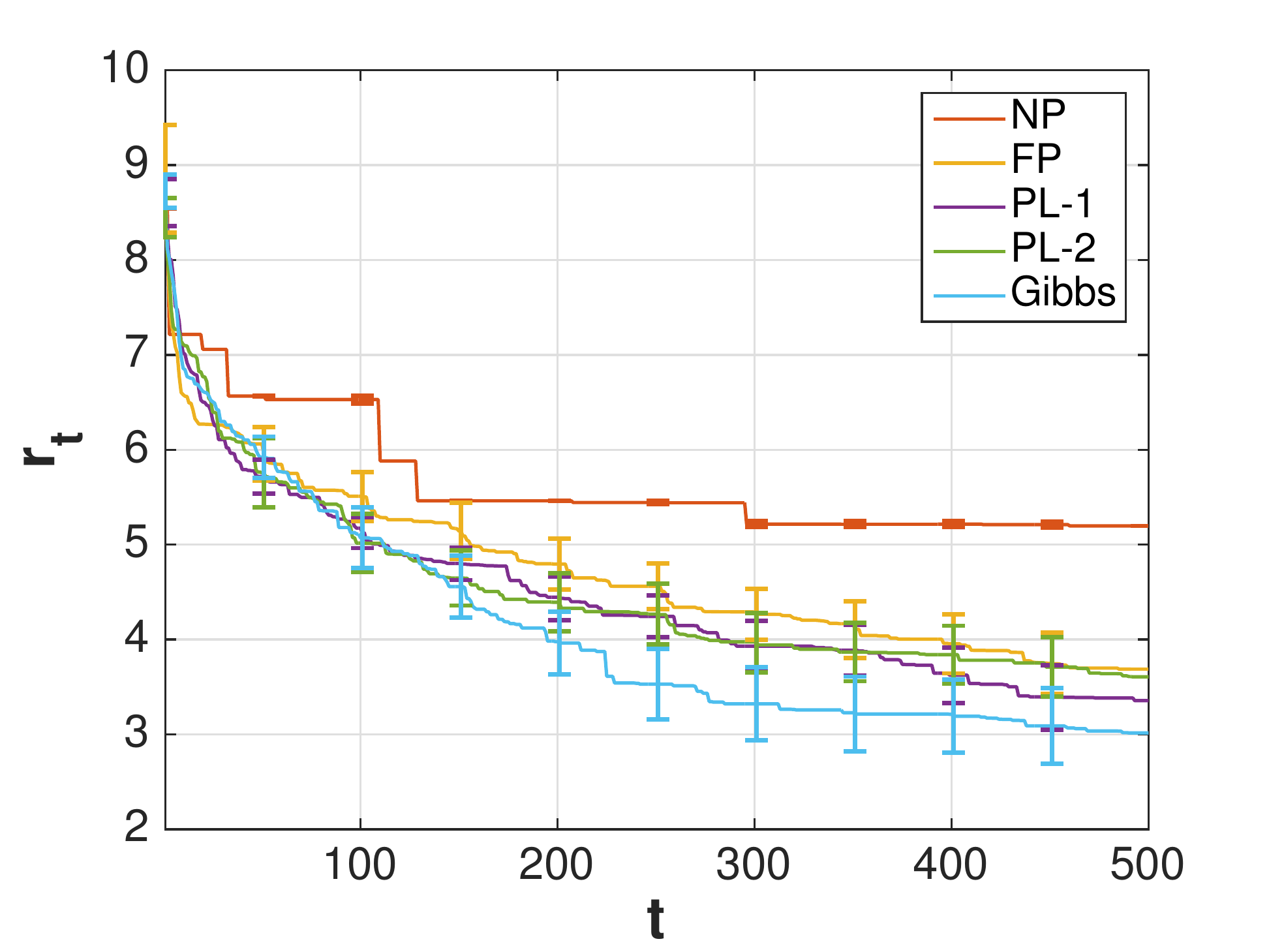}
\caption{Simple regret of tuning the 14 parameters for a robot pushing task. Learning decompositions with \texttt{Gibbs} is more effective than partial learning (\texttt{PL-1}, \texttt{PL-2}), no partitions (\texttt{NP}), or fully partitioned (\texttt{FP}). Learning decompositions with \texttt{Gibbs} helps BO to find a better point for this tuning task.}
\label{fig:seqrobot}
\end{figure}

Next, we test the decomposition learning algorithm on a real-world function, which returns the distance between a designated goal location and two objects being pushed by two robot hands, whose trajectory is determined by  14 parameters specifying the location, rotation, velocity, moving direction etc. This function is implemented with a physics engine, the Box2D simulator~\cite{box2d}. We use add-GP-UCB with different ways of setting the additive structure to tune the parameters for the robot hand so as to push the object closer to the goal. The regrets are shown in~Fig.~\ref{fig:seqrobot}. We observe that the performance of learning the decomposition with \texttt{Gibbs} dominates all existing alternatives including partial learning. Since the function we tested here is composed of the distance to two objects, there could be some underlying additive structure for this function in certain regions of the input space, e.g. when the two robots hands are relatively distant from each other so that one of the hands only impacts one of the objects. Hence, it is possible for \texttt{Gibbs} to learn a good underlying additive structure and perform effective BO with the structures it learned.

\subsection{Diverse Batch Sampling}
\begin{figure*}[h!]
\centering
\includegraphics[width=\textwidth]{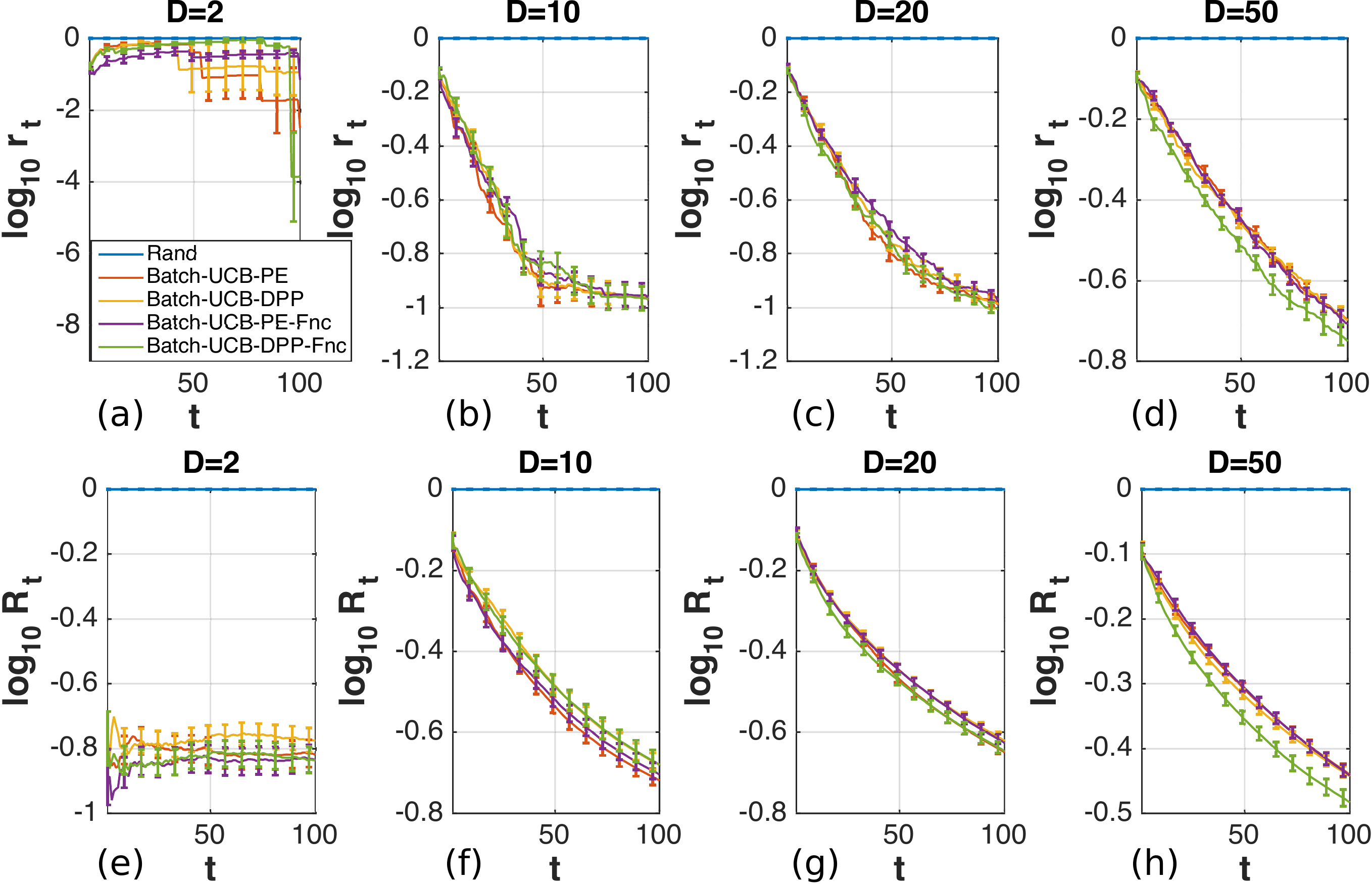}
\caption{Scaled simple regrets ($r_t$) and scaled averaged cumulative regrets ($R_t$) on synthetic functions with various dimensions when the ground truth decomposition is known. The batch sampling methods~(\texttt{Batch-UCB-PE}, \texttt{Batch-UCB-DPP}, \texttt{Batch-UCB-PE-Fnc} and \texttt{Batch-UCB-DPP-Fnc}) perform comparably well and outperform random sampling~(\texttt{Rand}) by a large gap.}
\label{fig:synthbatch}
\end{figure*}

Next, we probe the effectiveness of batch BO in high dimensions. In particular, we compare variants of the Add-UCB-DPP-BBO approach outlined in Section \ref{sec:batch}, and a baseline: %
\begin{itemize}
\item \texttt{Rand}: All batch points are chosen uniformly at random from $\cx$.
\item \texttt{Batch-UCB-*}: \texttt{*}$\in\{\texttt{PE},\texttt{DPP}\}$. All acquisition functions are UCB~(Eq.~\ref{eq:ucb}). Exploration is done via PE or DPP with posterior covariance kernels for each group. Combination is via sampling without replacement.
\item \texttt{*-Fnc}: \texttt{*}$\in\{\texttt{Batch-UCB-PE},\texttt{Batch-UCB-DPP}\}$. All quality functions are also UCB's, and combination is done by maximizing the quality functions.
\end{itemize}
A direct application of existing batch selection methods is very inefficient in the high-dimensional settings where they differ more, algorithmically, from our approach that exploits decompositions. Hence, we only compare to uniform sampling as a baseline.

\paragraph{Effectiveness}
We tested on $2$, $10$, $20$ and $50$-dimensional functions sampled the same way as in Section~\ref{sec:exp}; we assume the ground-truth decomposition of the feature space is known. Since \texttt{Rand} performs the worst, we show relative averaged cumulative regret and simple regret of all methods compared to \texttt{Rand} in Fig.~\ref{fig:synthbatch}. Results for absolute values of regrets are shown in the appendix. Each experiment was repeated for $20$ times. For all experiments, we set $\beta^m_t=|A_m|\log 2t$ and $B = 10$. All diverse batch sampling methods perform comparably well and far better than \texttt{Rand}, although there exist slight differences. While in lower dimensions ($D\in\{2,10\}$), \texttt{Batch-UCB-PE-Fnc} performs among the best, in higher dimensions ($D\in\{20,50\}$), \texttt{Batch-UCB-DPP-Fnc} performs better than (or comparable to) all other variants. We will see a larger performance gap in later real-world experiments, showing that biasing the combination towards higher quality functions while retaining diversity across the batch of samples provides a better exploration-exploitation trade-off.

\begin{figure}[h!]
\centering
\includegraphics[width=0.8\columnwidth]{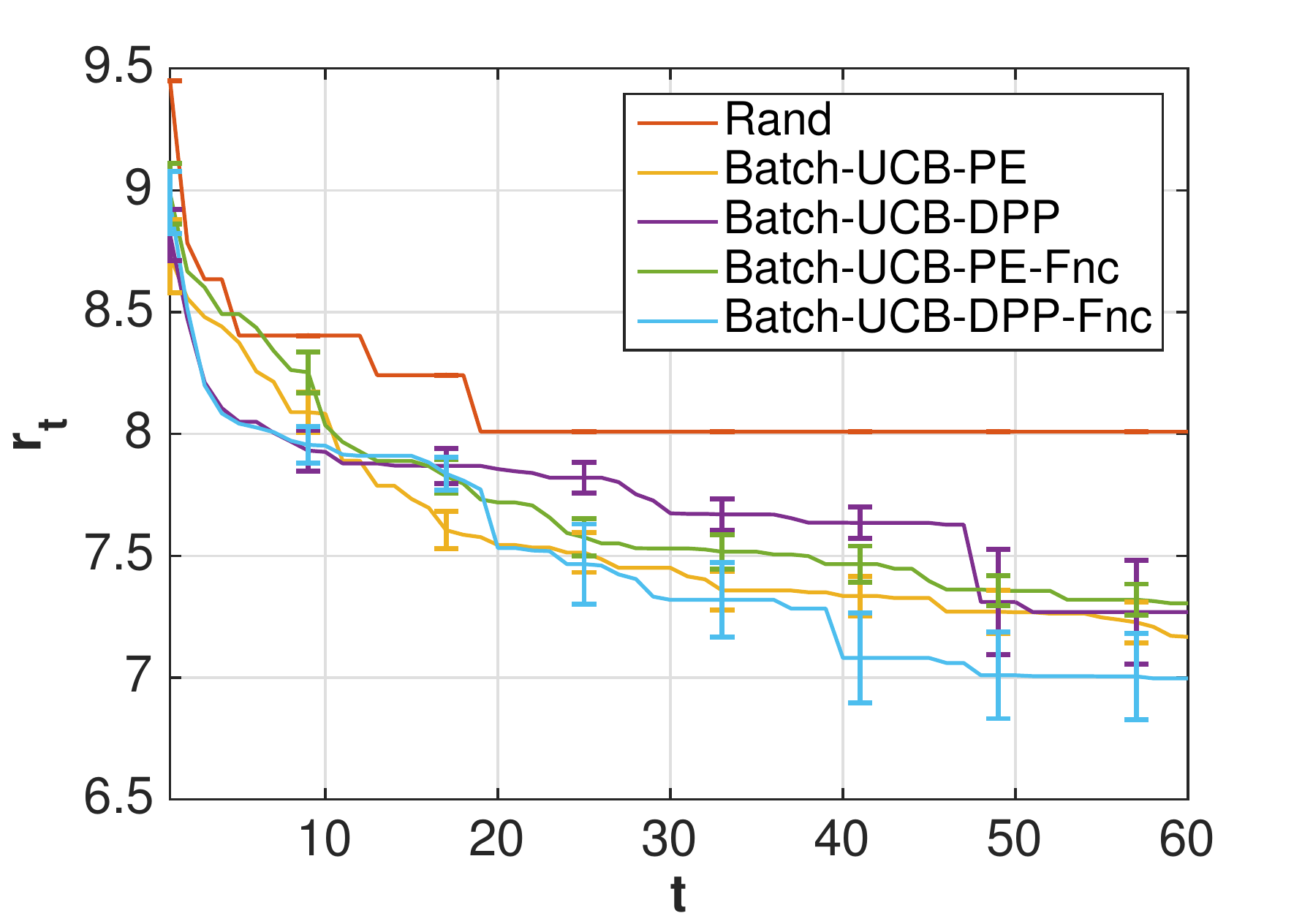}
\caption{The simple regrets ($r_t$) of batch sampling methods on \texttt{Walker} data where $B = 5$. Four diverse batch sampling methods~(\texttt{Batch-UCB-PE}, \texttt{Batch-UCB-DPP}, \texttt{Batch-UCB-PE-Fnc} and \texttt{Batch-UCB-DPP-Fnc}) outperform random sampling~(\texttt{Rand}) by a large gap. \texttt{Batch-UCB-DPP-Fnc} performs the best among the four diverse batch sampling methods. }
\label{fig:batch_walker_simple}
\end{figure}

For a real-data experiment, we tested the diverse batch sampling algorithms for BBO on the \texttt{Walker} function which returns the walking speed of a three-link planar bipedal walker implemented in Matlab~\cite{westervelt2007feedback}. We tune 25 parameters that may influence the walking speed, including 3 sets of 8 parameters for the ODE solver and 1 parameter specifying the initial velocity of the stance leg. We discretize each dimension into $40$ points, resulting in a function domain of $|\cx| = 40^{25}$. This size is very inefficient for existing batch sampling techniques. We learn the additive structure via Gibbs sampling and sample batches of size $B = 5$. To further improve efficiency, we limit the maximum size of each group to $2$. The regrets for all methods are shown in Fig.~\ref{fig:batch_walker_simple}. Again, all diverse batch sampling methods outperform \texttt{Rand} by a large gap. Moreover, \texttt{Batch-UCB-DPP-Fnc} is a bit better than other variants, suggesting that a selection by quality functions is useful. %

\paragraph{Batch Sizes}
Finally, we show how the batch size $B$ affects the performance of the proposed methods. We test the algorithms on the $14$-dimensional \texttt{Robot} dataset with $B\in\{5,10\}$. The regrets are shown in Fig.~\ref{fig:seqrobot}. With larger batches, the differences between the batch selection approaches become more pronounced. In both settings, \texttt{Batch-UCB-DPP-Fnc} performs a bit better than other variants, in particular with larger batch sizes. %

\begin{figure}[h!]
\centering
\includegraphics[width= \columnwidth]{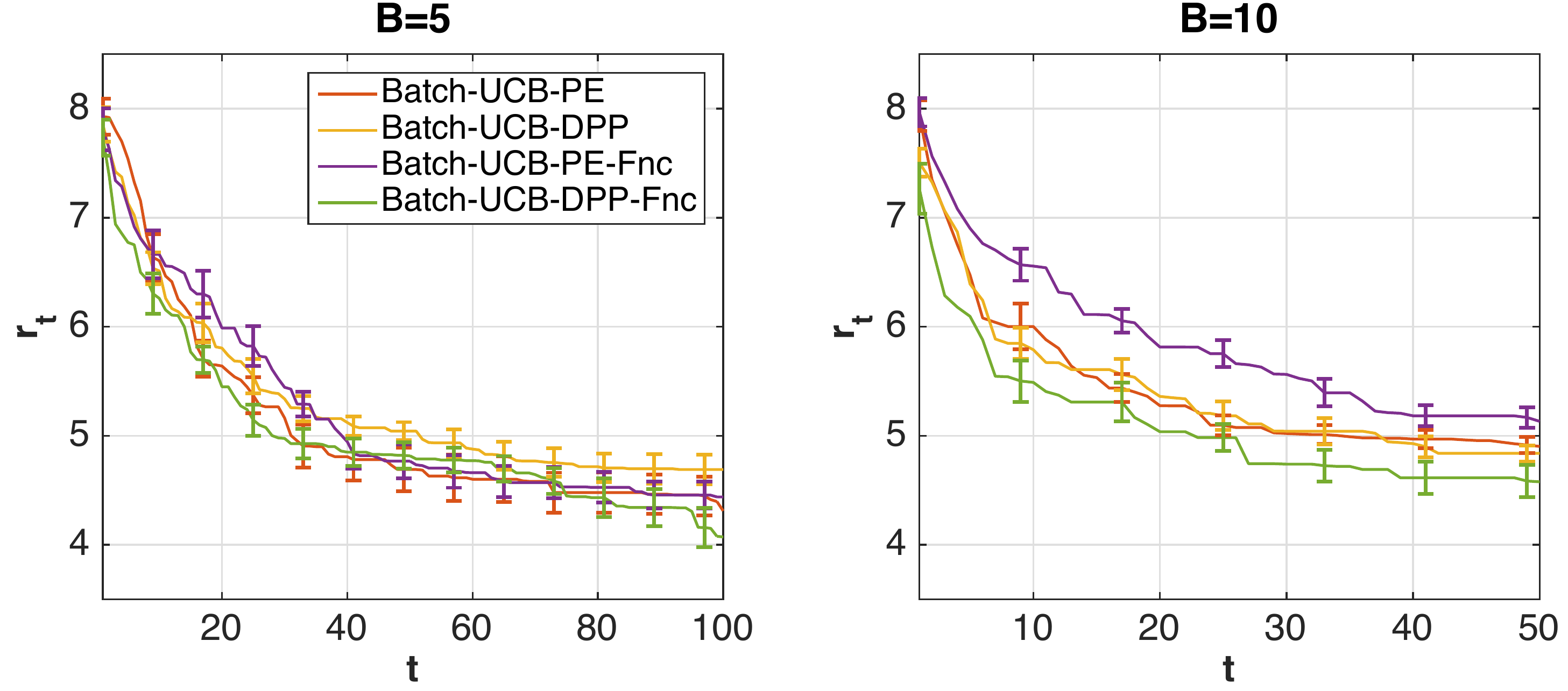}
\caption{Simple regret when tuning the 14 parameters of a robot pushing task with batch size 5 and 10. Learning decompositions with Gibbs sampling and diverse batch sampling are employed simultaneously. In general, \texttt{Batch-UCB-DPP-Fnc} performs a bit better than the other four diverse batch sampling variants. The gap increases with batch size.}
\label{fig:batch_robot_simple}
\end{figure}

\section{Conclusion}
In this paper, we propose two novel solutions for high dimensional BO: inferring latent structure, and combining it with batch Bayesian Optimization. The experimental results demonstrate that the proposed techniques are effective at optimizing high-dimensional black-box functions. Moreover, their gain over existing methods increases as the dimensionality of the input grows. We believe that these results have the potential to enable the increased use of Bayesian optimization for challenging black-box optimization problems in machine learning that typically involve a large number of parameters.

\section*{Acknowledgements}
We gratefully acknowledge support from NSF CAREER award 1553284, NSF grants 1420927 and 1523767, from ONR grant N00014-14-1-0486, and from ARO grant W911NF1410433.  We thank MIT Supercloud and the Lincoln Laboratory Supercomputing Center for providing computational resources. Any opinions, findings, and conclusions or recommendations expressed in this material are those of the authors and do not necessarily reflect the views of our sponsors.

\bibliography{refs}
\bibliographystyle{icml2017}
\newpage
\appendix

\section{Add-UCB-DPP-BBO Algorithm}

We present four variants of Add-UCB-DPP-BBO in Algorithm~\ref{app:algo:add_ucb_dpp_bbo}. The algorithm framework is general in that, one can plug in other acquisition and quality functions other than UCB to get different algorithms.

\begin{algorithm*}
\begin{algorithmic}
\STATE \textbf{Input}: $\cx$, $N_{init}$, $N_{cyc}$, $T$, $B$, $M$
\STATE Observe function values of $N_{init}$ points chosen randomly from $\cx$
\STATE Get the initial decomposition of feature space via Gibbs sampling and get corresponding $\cx_m$'s
\FOR{$t = 1$ to $T$}
	\IF{$(t\mod N_{cyc} = 0)$}
		\STATE Learn the decomposition via Gibbs sampling and get corresponding $\cx_m$'s
	\ENDIF
	\STATE Choose $x_0$ by maximizing UCB (acquisition function) for each group and combine them
	\FOR{$m=1$ to $M$}
    	\STATE Compute $(\calr_t^{(m)})^+$ and $\mathbf{K}^{(m)}_{(t-1)B+1}$
    	\STATE Sample $\{x_i^{(m)}\}_{i\in[B-1]} \subseteq \cx_m$ via PE or DPP with kernel $\mathbf{K}^{(m)}_{(t-1)B+1}$ 
	\ENDFOR
	\STATE Combine $\{x_i^{(m)}\}_{i\in[B-1], m\in[M]}$ either randomly or by maximizing UCB (quality function) without replacement to get $\{x_i\}_{i\in[B-1]}$ 
	\STATE Observe (noisy) function values for $\{x_i\}$ for $i\in\{0,\ldots,B-1\}$.
\ENDFOR
\end{algorithmic}
\caption{Add-UCB-DPP-BBO Variants}
\label{app:algo:add_ucb_dpp_bbo}
\end{algorithm*}
\section{Additional experiments}

\begin{table*}[h!]
\caption{Rand Index of the decompositions computed by Gibbs sampling.}
\label{tab:ri_alpha_gibbs}
\begin{center}
\begin{sc}
\begin{tabular}{l*{5}{c}}
\diagbox[width=3em]{$D$}{N} & 50 & 150 & 250 & 350 & 450 \\
\hline
5 & $0.85 \pm 0.20$ & $0.83 \pm 0.23$ & $0.71 \pm 0.18$ & $0.68 \pm 0.16$ & $0.66 \pm 0.18$  \\
10 & $0.78 \pm 0.06$ & $0.85 \pm 0.08$ & $0.86 \pm 0.10$ & $0.89 \pm 0.12$ & $0.95 \pm 0.06$  \\
20 & $0.88 \pm 0.02$ & $0.88 \pm 0.02$ & $0.89 \pm 0.02$ & $0.92 \pm 0.02$ & $0.95 \pm 0.04$  \\
50 & $0.95 \pm 0.01$ & $0.95 \pm 0.01$ & $0.95 \pm 0.01$ & $0.95 \pm 0.01$ & $0.95 \pm 0.01$ \\
100 & $0.98 \pm 0.00$ & $0.97 \pm 0.00$ & $0.97 \pm 0.00$ & $0.97 \pm 0.00$ & $0.97 \pm 0.00$ \\
\end{tabular}
\end{sc}
\end{center}
\vskip -0.3in
\end{table*}

\begin{table*}[h!]
\caption{Empirical posterior of any two dimensions correctly being grouped together by Gibbs sampling.}
\label{tab:learn_partition_together}
\begin{center}
\begin{sc}
\begin{tabular}{l*{6}{c}}
\diagbox[width=3em]{$d_x$}{N} & 50 & 150 & 250 & 350 & 450 \\
\hline
5 & $0.81 \pm 0.28$ & $0.91 \pm 0.19$ & $1.00 \pm 0.03$ & $0.97 \pm 0.08$ & $1.00 \pm 0.00$ & \\
10 & $0.21 \pm 0.13$ & $0.54 \pm 0.25$ & $0.68 \pm 0.25$ & $0.81 \pm 0.27$ & $0.93 \pm 0.15$ & \\
20 & $0.06 \pm 0.06$ & $0.11 \pm 0.08$ & $0.20 \pm 0.12$ & $0.43 \pm 0.17$ & $0.71 \pm 0.22$ & \\
50 & $0.02 \pm 0.03$ & $0.02 \pm 0.02$ & $0.03 \pm 0.03$ & $0.04 \pm 0.03$ & $0.06 \pm 0.04$ & \\
100 & $0.01 \pm 0.01$ & $0.01 \pm 0.01$ & $0.01 \pm 0.01$ & $0.01 \pm 0.01$ & $0.02 \pm 0.02$ & \\
\end{tabular}
\end{sc}
\end{center}
\vskip -0.1in
\end{table*}

\begin{table*}[h!]
\caption{Empirical posterior of any two dimensions correctly being separated by Gibbs sampling.}
\label{tab:learn_partition_separate}
\begin{center}
\begin{sc}
\begin{tabular}{l*{6}{c}}
\diagbox[width=3em]{$d_x$}{N} & 50 & 150 & 250 & 350 & 450 \\
\hline
2 & $0.30 \pm 0.46$ & $0.30 \pm 0.46$ & $0.90 \pm 0.30$ & $0.90 \pm 0.30$ & $1.00 \pm 0.00$ & \\
5 & $0.87 \pm 0.17$ & $0.80 \pm 0.27$ & $0.60 \pm 0.32$ & $0.55 \pm 0.29$ & $0.50 \pm 0.34$ & \\
10 & $0.88 \pm 0.05$ & $0.89 \pm 0.06$ & $0.89 \pm 0.07$ & $0.91 \pm 0.08$ & $0.94 \pm 0.07$ & \\
20 & $0.94 \pm 0.02$ & $0.94 \pm 0.02$ & $0.94 \pm 0.02$ & $0.95 \pm 0.02$ & $0.97 \pm 0.02$ & \\
50 & $0.98 \pm 0.00$ & $0.98 \pm 0.00$ & $0.98 \pm 0.01$ & $0.98 \pm 0.00$ & $0.98 \pm 0.01$ & \\
100 & $0.99 \pm 0.00$ & $0.99 \pm 0.00$ & $0.99 \pm 0.00$ & $0.99 \pm 0.00$ & $0.99 \pm 0.00$ & \\

\end{tabular}
\end{sc}
\end{center}
\vskip -0.1in
\end{table*}

In this section, we provide more details in our experiments. 
\subsection{Optimization of the Acquisition Functions}
We decompose the acquisition function into $M$ subacquisition
functions, one for each part, and optimize those separately. We
randomly sample 10000 points in the low dimensional
space, and then choose the one with the best value to start gradient descent
in the search space (i.e. the range of the box on $R^{|A_m|}$). In practice, we observe this approach optimizes low-dimensional ($< 5$ dimensions) functions very well. As the number of dimensions grows, the known difficulties of high dimensional
BO (and global nonconvex optimization) arise. 

\subsection{Effectiveness of Decomposition Learning}
\paragraph{Recovering Decompositions} In Table~\ref{tab:ri_alpha_gibbs}, Table~\ref{tab:learn_partition_together} and Table~\ref{tab:learn_partition_separate}, we show three quantities which may imply the quality of the learned decompositions.  The first quantity , reported in Table~\ref{tab:ri_alpha_gibbs}, is the Rand Index of the decompositions learned by Gibbs sampling, namely, $\frac{\sum_{i<j\leq D}\mathbbm{1}_{z^{g}_i\equiv z^{g}_j\wedge z_i\equiv z_j}+\sum_{i<j\leq D}\mathbbm{1}_{z^{g}_i\neq z^{g}_j\wedge z_i\neq z_j}}{\binom{D}{2}}$. The second quantity, reported in Table~\ref{tab:learn_partition_together}, is the probability of two dimensions being correctly grouped together by Gibbs sampling in each iteration of Gibbs sampling after the burn-in period, namely, $\frac{\sum_{i<j\leq D}\mathbbm{1}_{z^{g}_i\equiv z^{g}_j\wedge z_i\equiv z_j}}{\sum_{i<j\leq D}\mathbbm{1}_{z_i\equiv z_j}}$. The third quantity, reported in Table~\ref{tab:learn_partition_separate}, is the probability of two dimensions being correctly separated by Gibbs sampling in each iteration of Gibbs sampling after the burn-in period, namely, $\frac{\sum_{i<j\leq D}\mathbbm{1}_{z^{g}_i\neq z^{g}_j\wedge z_i\neq z_j}}{\sum_{i<j\leq D}\mathbbm{1}_{z_i\neq z_j}}$. 

\paragraph{Sensitivity Analysis for $\alpha$}
Empirically,
we found that the quality of the learned decompositions is not very sensitive to the
scale of $\alpha$ (see Table~\ref{tab:ri_alpha_gibbs}), because the log data likelihood plays a much more
important role than $\log(|A_m|+\alpha)$ when $\alpha$ is less than
the total number of dimensions. The reported results correspond to
alpha = 1 for all the partitions. 

\begin{table*}[h!]
\caption{Rand Index of the decompositions learned by Gibbs sampling for different values of $\alpha$.}
\label{tab:ri_alpha_gibbs}
\begin{center}
\begin{sc}
\begin{tabular}{l*{5}{c}}
\diagbox[width=3em]{$\alpha$}{N} & 50 & 150 & 250 & 350 & 450 \\
\hline

$0.2$ & $0.87811 \pm 0.019002$& $0.90126 \pm 0.022394$& $0.95284 \pm 0.047111$& $0.98811 \pm 0.02602$& $0.98811 \pm 0.026322$\\
$0.5$ & $0.88211 \pm 0.019893$& $0.90305 \pm 0.024574$& $0.95295 \pm 0.046232$& $0.98947 \pm 0.025872$& $0.99505 \pm 0.013881$\\
$1$ & $0.88211 \pm 0.016947$& $0.90326 \pm 0.024935$& $0.95305 \pm 0.043878$& $0.98558 \pm 0.034779$& $0.98053 \pm 0.035843$\\
$2$ & $0.88084 \pm 0.016972$& $0.9 \pm 0.023489$& $0.95463 \pm 0.042968$& $0.97989 \pm 0.038818$& $0.98832 \pm 0.023592$\\
$5$ & $0.88337 \pm 0.015784$& $0.90158 \pm 0.02203$& $0.96126 \pm 0.037045$& $0.98716 \pm 0.030949$& $0.99316 \pm 0.015491$\\

\end{tabular}
\end{sc}
\end{center}
\end{table*}

\begin{figure}
\centering
\includegraphics[width=0.5\textwidth]{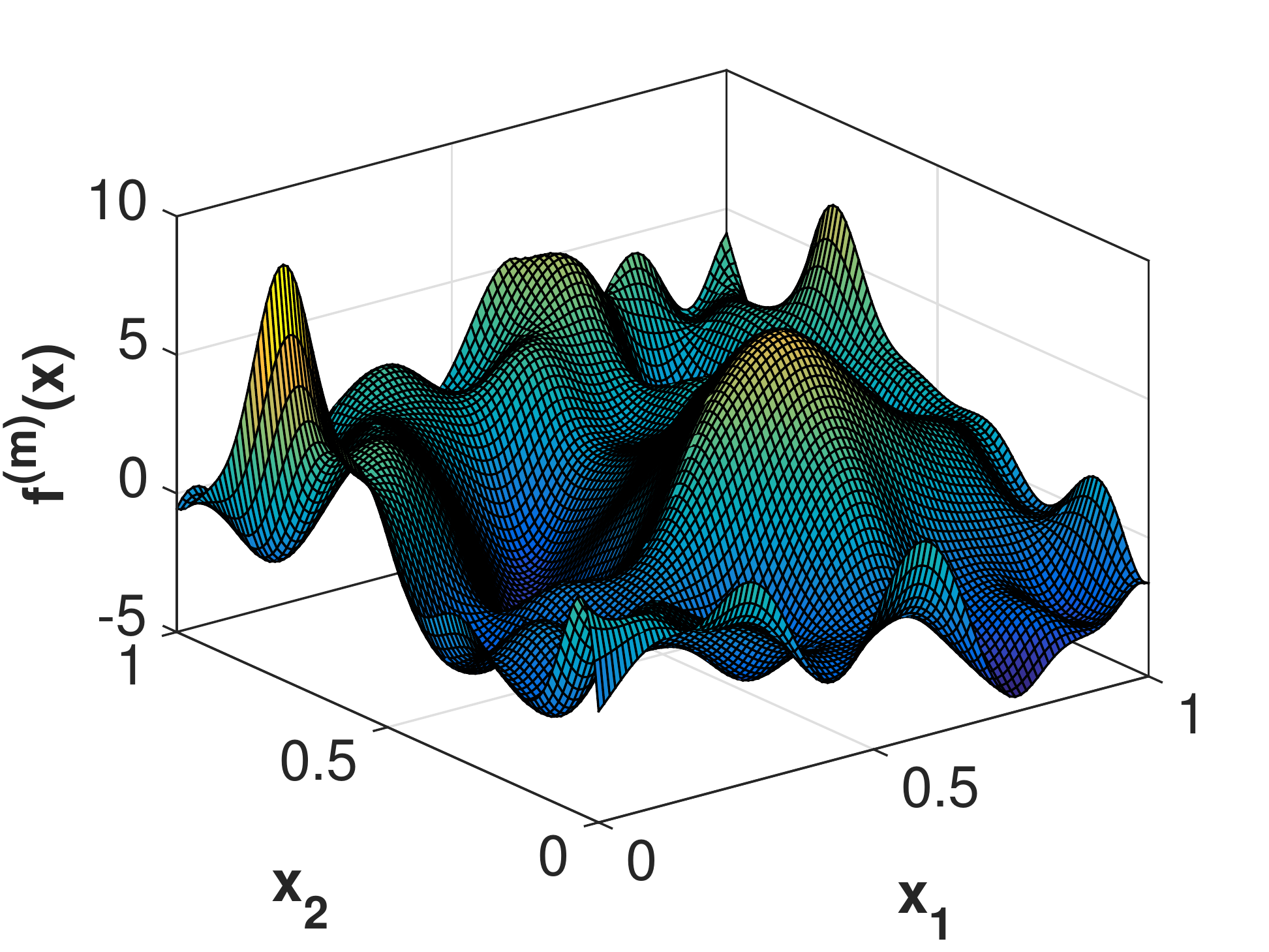}
\caption{An example of a 2 dimensional function component of the synthetic function.}
\label{fig:synth1example}
\end{figure}

\paragraph{BO for Synthetic Functions}  We show an example of a 2 dimensional function component in the additive synthetic function in Fig.~\ref{fig:synth1example}. Because of the numerous local maxima, it is very challenging to achieve the global optimum even for 2 dimensions, let alone maximizing an additive sum of them, only by  observing their sum. The full results of the simple and cumulative regrets for the synthetic functions comparing Add-GP-UCB with known additive structure (\texttt{Known}), no partitions (\texttt{NP}), fully partitioned with one dimension for each group (\texttt{FP})  and the following methods of learning partition: Gibbs sampling (\texttt{Gibbs}), random sampling the same number of partitions sampled by \texttt{Gibbs} and select the one with the highest data likelihood (\texttt{PL-1}), random sampling 5 partitions and select the one with the highest data likelihood (\texttt{PL-2}) are shown in Fig.~\ref{fig:synth1}. The learning was done every 50 iterations, starting from the first iteration. For $D=20,30$, it is quite obvious that when a new partition is learned from the newly observed data (e.g. at iteration 100 and 150), the simple regret gets a boost.

\begin{figure}
\centering
\includegraphics[width=\columnwidth]{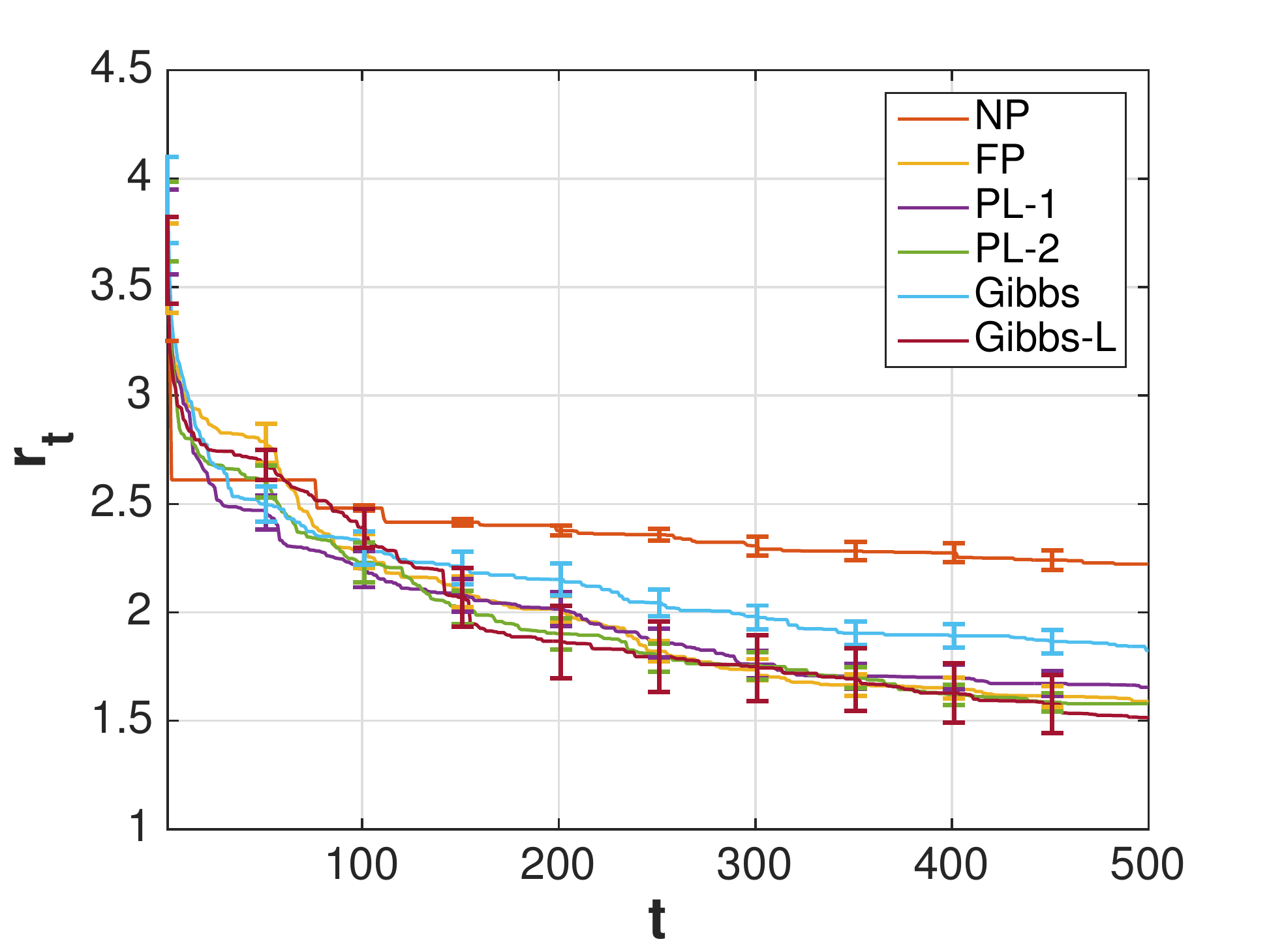}
\caption{Simple regret of tuning the 25 parameters for optimizing the walking speed of a bipedal robot. We use the vanilla Gibbs sampling algorithm (\texttt{Gibbs}) and a Gibbs sampling algorithm with partition size limit set to be 2 (\texttt{Gibbs-L}) to compare with partial learning (\texttt{PL-1}, \texttt{PL-2}), no partitions (\texttt{NP}), and fully partitioned (\texttt{FP}). \texttt{Gibbs-L} performed slightly better than \texttt{PL-2} and \texttt{FP}. This function does not have an additive structure, and as a result, \texttt{Gibbs} does not perform well for this function because the sizes of the groups it learned tend to be large .}
\label{fig:seqwalker}
\end{figure}

\begin{figure*}
\centering
\includegraphics[width=0.9\textwidth]{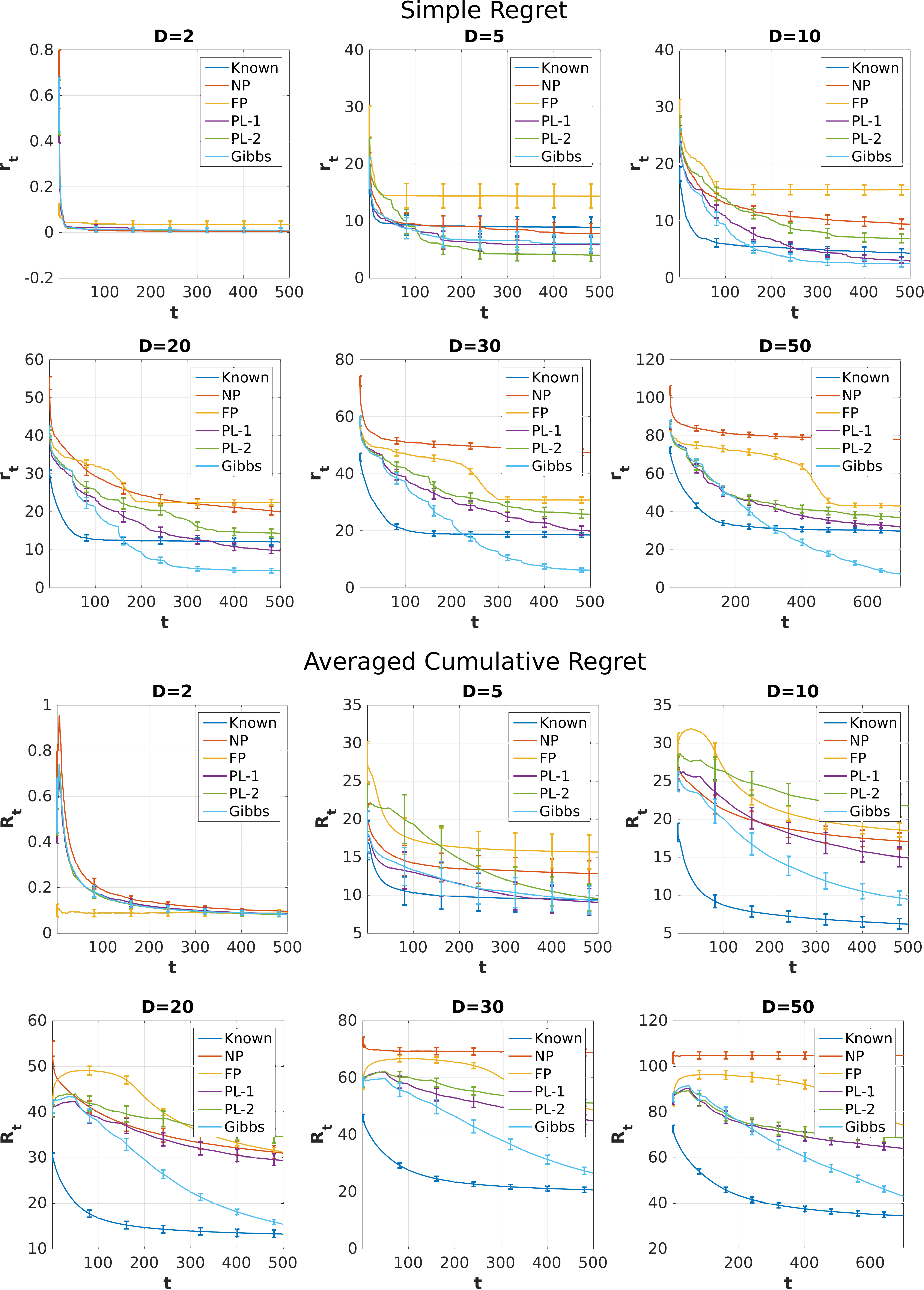}
\caption{The simple regrets ($r_t$) and the averaged cumulative regrets ($R_t$) and  for \texttt{Known} (ground truth partition is given),  \texttt{Gibbs} (using Gibbs sampling to learn the partition), \texttt{PL-1} (randomly sample the same number of partitions sampled by \texttt{Gibbs} and select the one with highest data likelihood), PL-2 (randomly sample 5 partitions and select the one with highest data likelihood), \texttt{FP} (fully partitioned, each group with one dimension) and \texttt{NP} (no partition) on 10, 20, 50 dimensional functions. \texttt{Gibbs} achieved comparable results to \texttt{Known}. Comparing \texttt{PL-1} and \texttt{PL-2} we can see that sampling more partitions did help to find a better partition. But a more principled way of learning partition using \texttt{Gibbs} can achieve much better performance than \texttt{PL-1} and \texttt{PL-2}.}
\label{fig:synth1}
\end{figure*}

\paragraph{BO for Real-world functions} 

In addition to be 14 parameter robot pushing task, we tested on the walker function which returns the walking speed of a three-link planar bipedal walker implemented in Matlab~\cite{westervelt2007feedback}. We tune 25 parameters that may influence the walking speed, including 3 sets of 8 parameters for the ODE solver and 1 parameter specifying the initial velocity of the stance leg. To our knowledge, this function does not have an additive structure. The regrets of each decomposition learning methods are shown in Fig.~\ref{fig:seqwalker}. In addition to \texttt{Gibbs}, we test learning decomposition via constrained Gibbs sampling (\texttt{Gibbs-L}), where the maximum size of each group of dimensions does not exceed 2. Because the function does not have additive structure, \texttt{Gibbs} performed poorly since it groups together many dimensions of the input. As a result, its performance is similar to that of no partition (\texttt{NP}). However, \texttt{Gibbs-L} appears to learn a good decomposition with the group size limit, and manages to achieve a slightly lower regret than other methods. \texttt{Gibbs}, \texttt{PL-1}, \texttt{PL-2} and \texttt{FP} all performed relatively well in for this function, indicating that using the additive structure may benefit the BO procedure even if the function itself is not additive.

\subsection{Diverse Batch Sampling}
\begin{figure*}[h!]
\centering
\includegraphics[width=.9\textwidth]{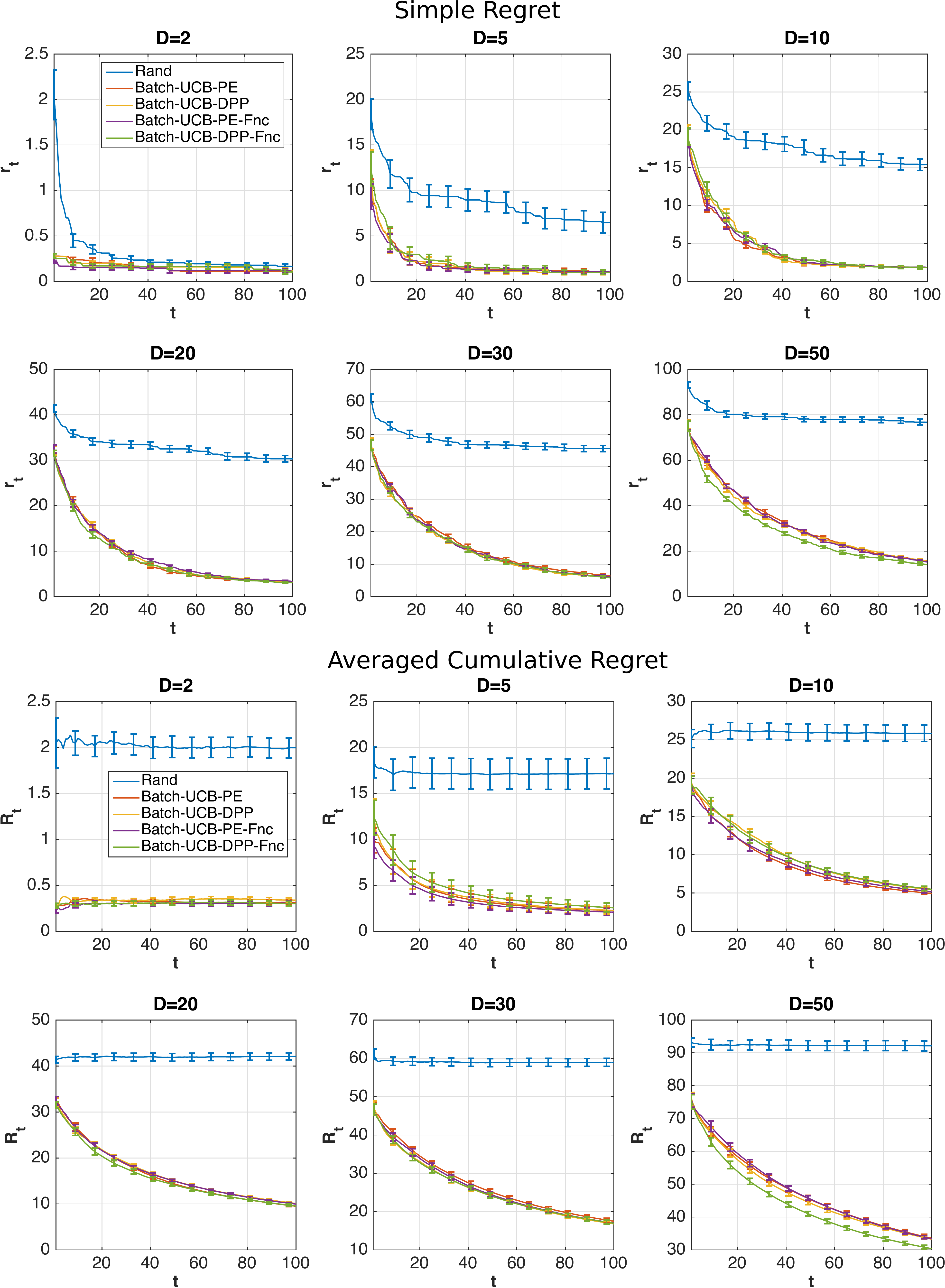}
\caption{The simple regrets ($r_t$) and the averaged cumulative regrets ($R_t$) on synthetic functions with various dimensions when the ground truth partition is known. Four batch sampling methods~(\texttt{Batch-UCB-PE}, \texttt{Batch-UCB-DPP}, \texttt{Batch-UCB-PE-Fnc} and \texttt{Batch-UCB-DPP-Fnc}) perform comparably well and outperform random sampling~(\texttt{Rand}) by a large gap.}
\label{fig:batch_synth_simple}
\end{figure*}

In Fig.~\ref{fig:batch_synth_simple}, we show the full results of the simple and the cumulative regrets on the synthetic functions described in Section 5.2 of the paper.

\hide{
Fig.~\ref{fig:batch_synth_improve} shows the improvement of \texttt{Batch-UCB-DPP-Fnc} over \texttt{Rand} in terms of regrets. The improvement in general grows with $d$, indicating that the diverse batch sampling in higher dimensional settings may benefit more.

\begin{figure}[h!]
\centering
\includegraphics[width=0.4\textwidth]{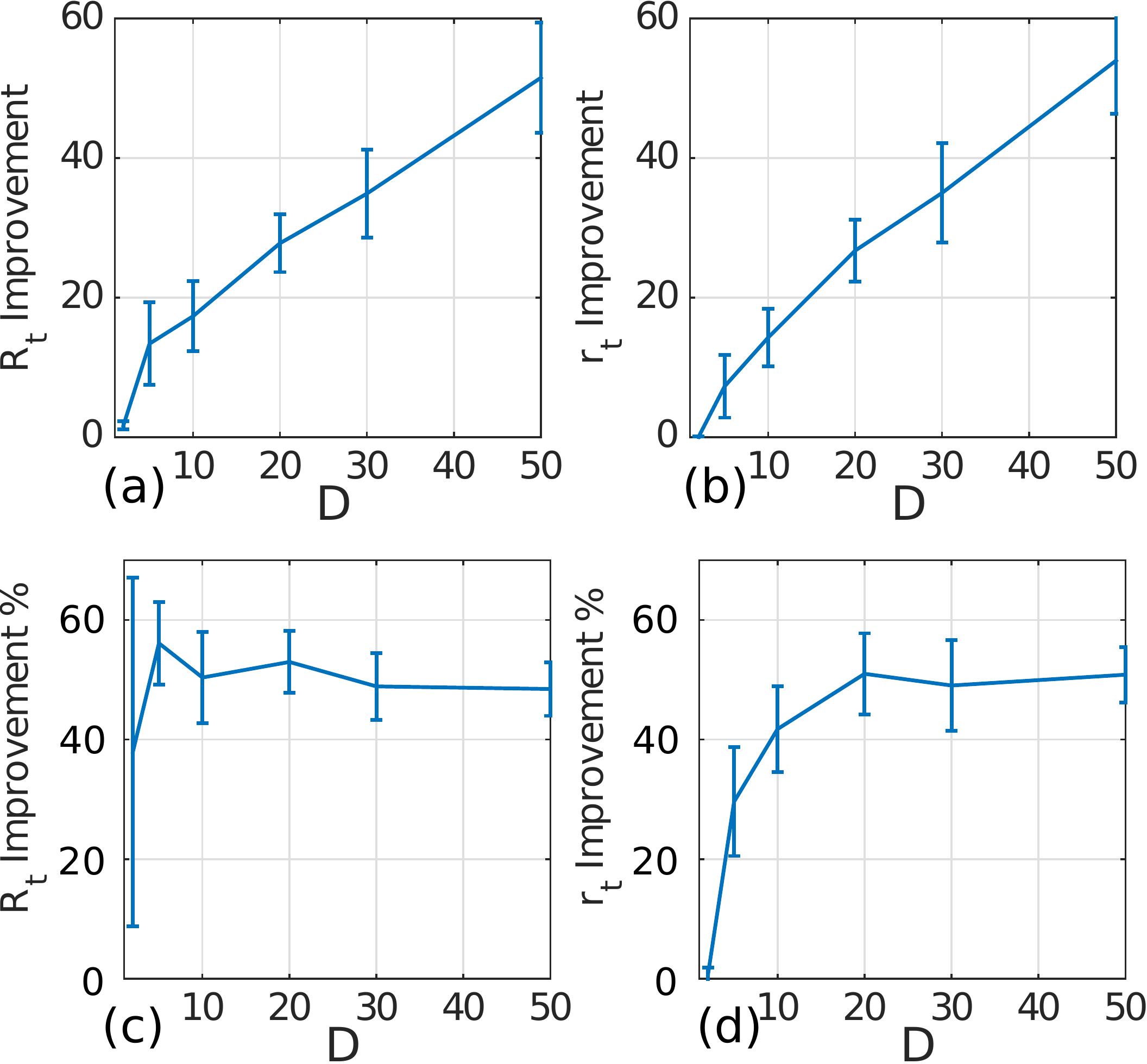}
\caption{Improvement made efficient diverse batch sampling method \texttt{Batch-UCB-DPP-Fnc} over \texttt{Rand}. (a) averaged cumulative regret; (b) simple regret. (c) averaged cumulative regret normalized by function maximum; (d) simple regret normalized by function maximum. Using diverse batch sampling continues to outperform BBO with random exploration.}
\label{fig:batch_synth_improve}
\end{figure}
}
\end{document}